\documentclass[conference]{IEEEtran}
\IEEEoverridecommandlockouts
\usepackage{cite}
\usepackage{amsmath,amssymb,amsfonts}
\usepackage{algorithmic}
\usepackage{graphicx}
\usepackage{textcomp}
\usepackage{xcolor}

\usepackage{multirow}
\usepackage{makecell}
\usepackage{booktabs}
\usepackage{tabularx}
\usepackage{tikz}
\usepackage{xspace}
\usepackage{float}
\usepackage{url}
\usepackage{hyperref}

\newcolumntype{b}{X}
\newcolumntype{s}{>{\hsize=.5\hsize}X}

\def\checkmark{\tikz\fill[scale=0.3](0,.35) -- (.25,0) -- (1,.7) -- (.25,.15) -- cycle;} 

\makeatletter
\DeclareRobustCommand\onedot{\futurelet\@let@token\@onedot}
\def\@onedot{\ifx\@let@token.\else.\null\fi\xspace}

\def\eg{\emph{e.g}\onedot}

\def\etal{\emph{et al}\onedot}
\makeatother

\def\BibTeX{{\rm B\kern-.05em{\sc i\kern-.025em b}\kern-.08em
    T\kern-.1667em\lower.7ex\hbox{E}\kern-.125emX}}
\begin{document}

\title{Aerial-Ground Person Re-ID\\
}

\author{\IEEEauthorblockN{Huy Nguyen, Kien Nguyen,  Sridha Sridharan, and Clinton Fookes}
\IEEEauthorblockA{\textit{School of Electrical Engineering and Robotics} \\
\textit{Queensland University of Technology}\\
Brisbane, Australia 4000 \\
thanhnhathuy.nguyen@hdr.qut.edu.au, \{k.nguyenthanh, s.sridharan, c.fookes\}@qut.edu.au}
}

\maketitle

\begin{abstract}
Person re-ID matches persons across multiple non-overlapping cameras. Despite the increasing deployment of airborne platforms in surveillance, current existing person re-ID benchmarks' focus is on ground-ground matching and very limited efforts on aerial-aerial matching. We propose a new benchmark dataset - AG-ReID, which performs person re-ID matching in a new setting: across aerial and ground cameras. Our dataset contains 21,983 images of 388 identities and 15 soft attributes for each identity. The data was collected by a UAV flying at altitudes between 15 to 45 meters and a ground-based CCTV camera on a university campus. Our dataset presents a novel elevated-viewpoint challenge for person re-ID due to the significant difference in person appearance across these cameras. We propose an explainable algorithm to guide the person re-ID model's training with soft attributes to address this challenge. Experiments demonstrate the efficacy of our method on the aerial-ground person re-ID task. The dataset will be published and the baseline codes will be open-sourced at \href{https://github.com/huynguyen792/AG-ReID}{https://github.com/huynguyen792/AG-ReID} to facilitate research in this area. 
\end{abstract}


\section{Introduction}
\begin{figure} 
	\centering
	\includegraphics[width=0.49\columnwidth ]{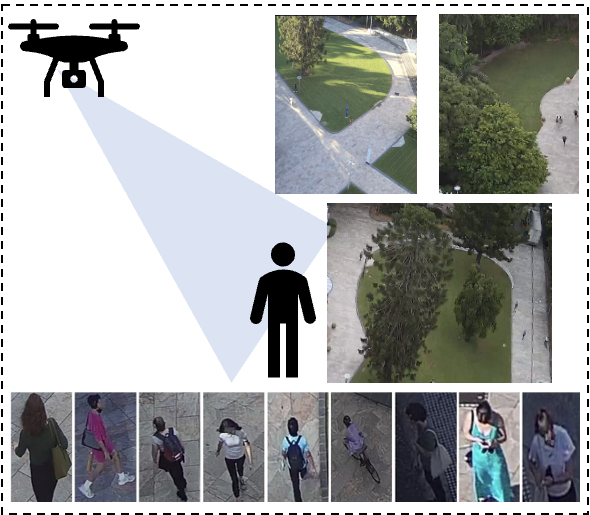}
	\includegraphics[width=0.49\columnwidth ]{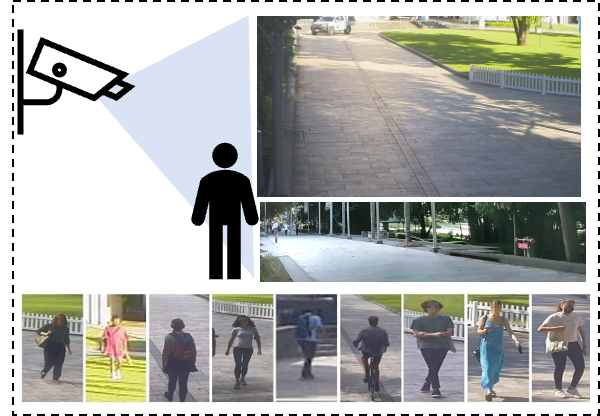} \\
	\small  (a) Aerial camera \hspace{0.25\columnwidth} (b) Ground camera 
	\caption{Significant difference between elevated and horizontal viewpoint due to cameras' diversity.}
	\label{fig:perspective}
\end{figure}

Aerial surveillance recently attracted researchers due to the advancement of  drones and cameras \cite{Li2021UAVHumanAL, AerialSurveillance, Zhu2020VisionMD} enabling new settings for person re-ID in aerial imagery. 
{Current research in aerial person re-identification is limited, primarily focusing on matching aerial images, with only one study by Schumann et al. \cite{Schumann2017PersonRA} exploring aerial-ground ReID, emphasizing the need for a large-scale, realistic dataset to advance this area.} This paper presents a unique collection of Aerial and Ground visual data called AG-ReID as the first public dataset for developing aerial-ground person re-ID. The AG-ReID dataset is collected with aerial and ground images of pedestrians on a university campus. The combination of an aerial platform (UAV/drone) and a fixed ground-based CCTV camera illustrates realistic real-world surveillance scenarios where resources on the ground and in the air are deployed collaboratively for the surveillance task. A DJI XT2 drone captures aerial images with a 24mm lens flying at multiple altitudes from 15 to 45 meters. The unique high-flying altitudes of aerial cameras make persons appear differently in viewpoints, poses and resolution compared to the images of the same person viewed from ground cameras. Therefore, re-identifying persons across aerial-ground setting present some unique and novel challenges in matching. Exemplar images collected in our AG-ReID dataset can be seen in Fig. \ref{fig:perspective}. Our dataset also characterizes each target person with 15 soft-biometric labels for attribute recognition and improving person re-ID. 

Learning across aerial-ground cameras has unique challenges compared to learning with ground-ground and aerial-aerial cameras. The critical challenge is the significant differences in person perspectives where persons appearing in aerial images have a higher elevated viewpoint than ground images, making matching difficult. For this paper, we propose an explainable approach to deal with these challenges of the aerial-ground person re-ID task based on soft-biometric attributes. The approach contains two streams, a transformer-based person re-ID model stream and an explainable person re-ID stream. We explore critical attributes contributing to the similarities between two persons where the person's perspective is different. A key point in the explainable approach is that it enables us to visualize the attributes map contributing to the differences between the images of the same person from different ground-aerial cameras. This enables us to explain the person-matching result since people with similar attributes are more likely to be the same person compared to people with similar visual appearances but different attributes. Our proposed approach enables us to explore and develop explainable methods to learn across aerial and ground person re-ID effectively.

We summarize below the four main contributions, including (i) a new aerial-ground person re-ID dataset named AG-ReID, (ii) a two-stream person re-ID model with an explainable approach, (iii) extensive experimental results, and (iv) public release of the AG-ReID dataset and the baseline codes for the two-stream approach.

\section{Related Work}


In this section, we examine and compare existing person re-ID datasets and discuss the explainable trend in person re-ID. 
     	          
\vspace{3px}
\noindent\textbf{Datasets for Person Re-ID}

\vspace{2px}
\noindent \textit{Ground-Ground Person Re-ID datasets}: This is the most popular setting, where person images across multiple ground cameras are matched. Two popular datasets are Market-1501 \cite{Zheng2015ScalablePR} and DukeMTMC-reID \cite{Gou2017DukeMTMC4ReIDAL}. Introduced in 2015, the Market-1501 dataset has 1,501 person identities, and a total of 32,668 person images \cite{Zheng2015ScalablePR}. Compared to Market-1501, DukeMTMC-reID, introduced by \cite{Gou2017DukeMTMC4ReIDAL}, has more pedestrian images (36,411) with fewer identities (1,404). 

\vspace{2px}                     
\noindent \textit{Aerial-Aerial Person Re-ID datasets}: This setting matches person images captured across multiple drones. Two datasets that exist in the literature are PRAI-1581 \cite{Zhang2021PersonRI} and UAV-Human \cite{Li2021UAVHumanAL}. Presented by \cite{Zhang2021PersonRI} in 2020, the PRAI-1581 contains 1,581 subjects and 39,461 images in total. UAV-Human was introduced by Li \etal \cite{Li2021UAVHumanAL} in 2021. UAV-Human was collected by one UAV, flying over three months at 2 to 8 meters in various areas and times (day and night). The dataset can be used for multiple surveillance tasks; however here we only review its person re-ID task. There are, in total, 1,144 identities and 41,290 person images for the aerial-aerial person re-ID task. 
	      	      	          
\vspace{2px}
\noindent \textit{Aerial-Ground Person Re-ID dataset}: To our knowledge, no research has been performed on a large dataset that is accessible to the public for re-identifying persons across aerial and ground cameras. Our work will be the first to contribute to this gap.

\vspace{3px}
\noindent\textbf{Explainable Person Re-ID} Due to the black box-ness of CNN models \cite{Somers2022BodyPR, Wang2021FeatureEA}, there is a trend in developing explainable approaches to perform person re-ID. Previous works attempt to explain CNN models by visualizing salient maps \cite{Selvaraju2019GradCAMVE}, distilling knowledge \cite{Chen2019ExplainingNN} or learning with a decision tree \cite{Zhang2019InterpretingCV}. In the explainable person re-ID domain, the works of \cite{Yang2019TowardsRF, Zhang2020RelationAwareGA} investigate attention learning as attempts to explain the person re-ID model's prediction. Chen \etal \cite{Chen2021ExplainablePR} proposed an interpreter for the person re-ID model using attributes. Although this work achieves high performance, their work focuses only on the ground-ground setting. In the aerial setting, there does not exist any explainable approach. The requirement of explainability is even more critical in the aerial-ground setting as a person's appearance could be significantly different. For this reason, as the first attempt to deal with the aerial-ground challenges, we develop an explainable person re-ID approach in the aerial-ground setting. 

\begin{figure}
    \centering
    \includegraphics[width=0.9 \columnwidth]{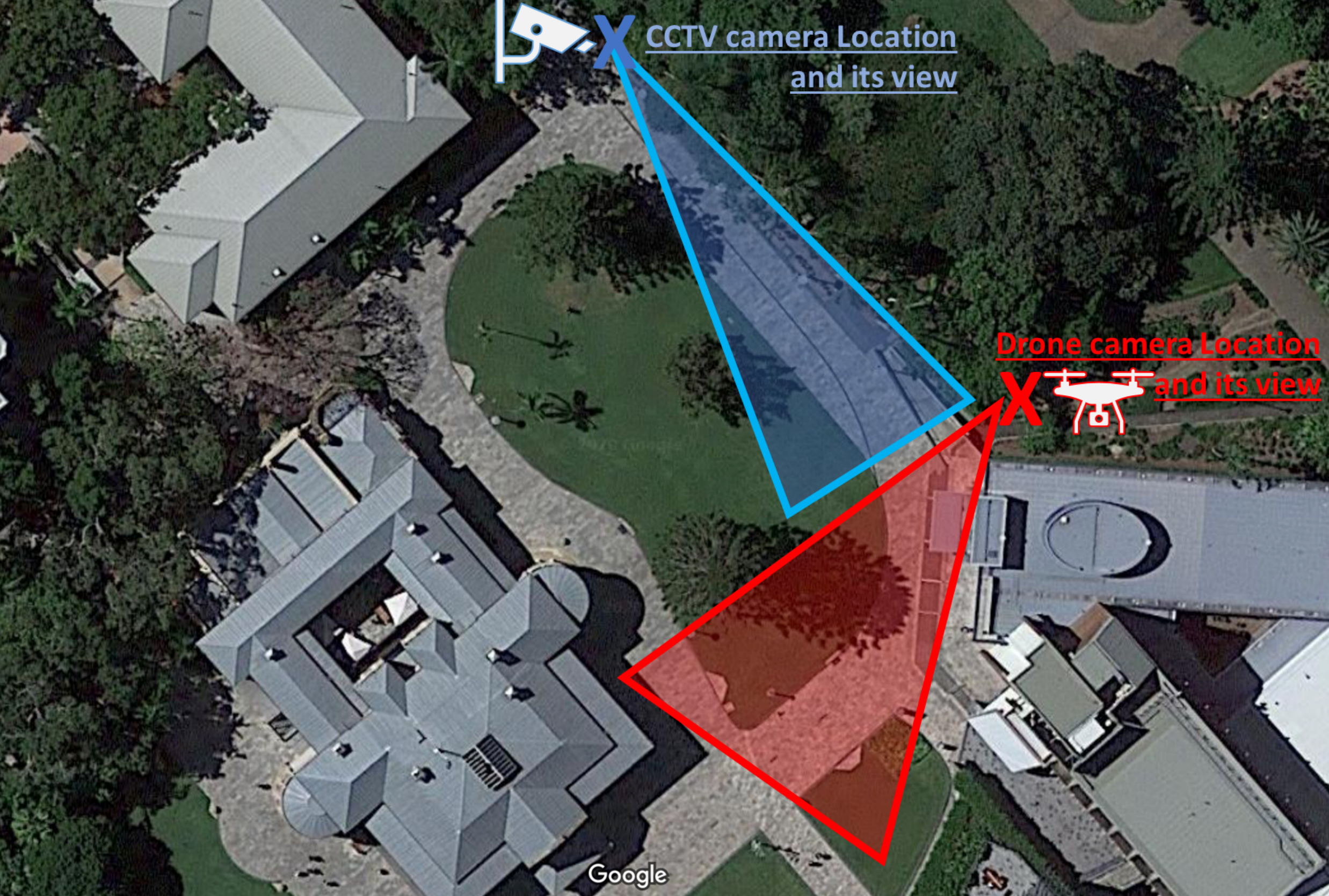}
    \caption{Data Collection Area: CCTV camera captures ground data in the blue area and UAV camera captures aerial data from the red area. Blue and red areas are non-overlapping. }
    \label{fig:DataCollectionArea}
\end{figure}

\section{AG-ReID Dataset}

\begin{table*}
 \fontsize{8}{10}\selectfont
	\caption{Comparison of our AG-ReID dataset with popular person re-ID datasets. As there is no public dataset in this new aerial-ground setting, we compare our dataset with existing public ground-ground and aerial-aerial datasets.}
	\begin{tabularx}{\textwidth}{XXXXXXX} 
		\hline
		\multicolumn{2}{c}{\multirow{2}{*}{\textbf{Dataset Attributes}}}                                            
		  & \multicolumn{2}{c}{\textbf{Ground-Ground }} & \multicolumn{2}{c}{\textbf{Aerial-Aerial }} & \multicolumn{1}{c}{\textbf{Aerial-Ground     }} \\ 
		\cline{3-7}
										
		\multicolumn{2}{c}{}                                  & \multicolumn{1}{c}{Market-1501} 
		&     \multicolumn{1}{c}{DukeMTMC-reID }         
		  & \multicolumn{1}{c}{PRAI-1581  }             & \multicolumn{1}{c}{UAV-Human }              & \multicolumn{1}{c}{AG-ReID (ours)}               
							
		\\ \hline
							
		\multirow{4}{*}{{Person Re-ID  }       }                             
		&        \multicolumn{1}{l}{\# Annotated IDs  }                    
		& \multicolumn{1}{c}{1,501  }                            
		&   \multicolumn{1}{c}{1,404   }             
		&  \multicolumn{1}{c}{1,581}    
		&  \multicolumn{1}{c}{1,144  }                                                 
		&    \multicolumn{1}{c}{388 }                                                  
		\\ 
		\cline{2-7}

		&        \multicolumn{1}{l}{\# Image Samples with IDs  }            
		&      \multicolumn{1}{c}{32,668    }                          
		&      \multicolumn{1}{c}{36,411   }          
		&      \multicolumn{1}{c}{39,461 }                   
		&  \multicolumn{1}{c}{41,290}                                         
		&     \multicolumn{1}{c}{21,983 }                                                     \\ 
		\cline{2-7}

								 
		& \# Attributes                              
		& \multicolumn{1}{c}{$\times$  }                       
		& \multicolumn{1}{c}{$\times$  }           
		& \multicolumn{1}{c}{$\times$  }            
		& \multicolumn{1}{c}{7  }                                    
		&   \multicolumn{1}{c}{15}
		\\ 
		\cline{2-7}
									
		& \multicolumn{1}{l}{\# Image Samples with Attributes} 
		& \multicolumn{1}{c}{$\times$ }            
		& \multicolumn{1}{c}{$\times$ } 
		& \multicolumn{1}{c}{ $\times$ } 
		& \multicolumn{1}{c}{{22,263}}
		& \multicolumn{1}{c}{21,983}
		\\ 	\hline

		\multicolumn{2}{c}{Diverse Backgrounds}               & \multicolumn{1}{c}{$\times$}   
		& \multicolumn{1}{c}{$\times$   }           
		& \multicolumn{1}{c}{$\times$   }           
		& \multicolumn{1}{c}{\textbf{$\checkmark$}  }                              
		& \multicolumn{1}{c}{\checkmark  }                                           \\ 
		\hline

		\multicolumn{2}{c}{Occlusion}                         & \multicolumn{1}{c}{$\times$}   
		& \multicolumn{1}{c}{$\times$   }           
		& \multicolumn{1}{c}{$\checkmark$    }      
		& \multicolumn{1}{c}{\textbf{$\checkmark$}}                                
		& \multicolumn{1}{c}{$\checkmark$ }                                           \\ 
		\hline

		\multicolumn{2}{c}{Camera Views}                      & \multicolumn{1}{c}{fixed}      
		& \multicolumn{1}{c}{fixed }              
		& \multicolumn{1}{c}{mobile}              
		& \multicolumn{1}{c}{mobile  }                                            
		& \multicolumn{1}{c}{fixed \& mobile}                                                \\ 
		\hline

		\multicolumn{2}{c}{Platforms}                      & \multicolumn{1}{c}{CCTV}      
		& \multicolumn{1}{c}{CCTV        }       
		& \multicolumn{1}{c}{UAV         }       
		& \multicolumn{1}{c}{UAV         }                                    
		& \multicolumn{1}{c}{UAV \& CCTV   }                                             \\ 
		\hline
									
		\multicolumn{2}{c}{Flying Altitude}                      & \multicolumn{1}{c}{$<10m$}      
		& \multicolumn{1}{c}{$<10m$    }
		& \multicolumn{1}{c}{$20\sim60m$    }           
		& \multicolumn{1}{c}{~ $2\sim8m$    }                                           
		& \multicolumn{1}{c}{~ $15\sim45m$  }                                               \\ 
		\hline
									
		\multicolumn{2}{c}{\#UAVs}                      & \multicolumn{1}{c}{0}      
		& \multicolumn{1}{c}{~ 0    }         
		& \multicolumn{1}{c}{~ 2   }          
		& \multicolumn{1}{c}{~ 1   }                                          
		& \multicolumn{1}{c}{~ 1    }                                           
		\\ 
		\hline

	\end{tabularx}
	\label{tab:compare_statics}
\end{table*}

\subsection{Dataset Description}
\begin{figure}[H]
					
	\centering
					  
	\begin{tabular}{@{}c@{}}
		\includegraphics[width=0.7\columnwidth, height = 1cm]{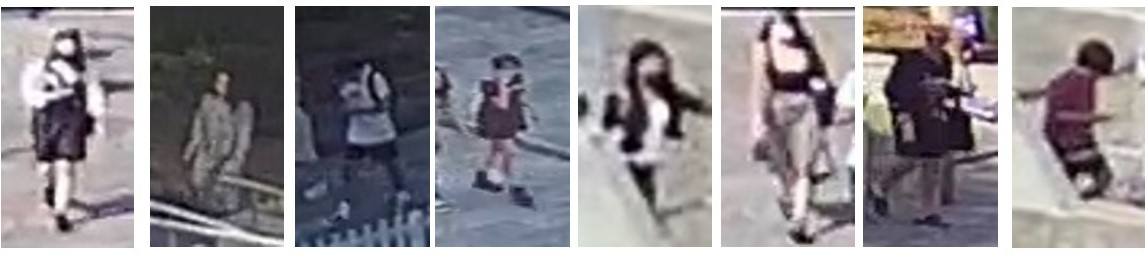} \\
		\small (a) Low Resolution                                                  
	\end{tabular}
					
	\begin{tabular}{@{}c@{}}
		\includegraphics[width=0.7\columnwidth, height = 1cm]{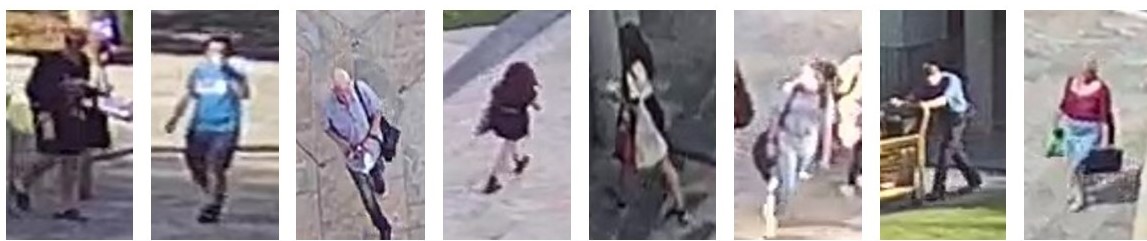} \\
		\small (b) Blur                                                  
	\end{tabular}
					
	\begin{tabular}{@{}c@{}}
		\includegraphics[width=0.7\columnwidth, height = 1cm]{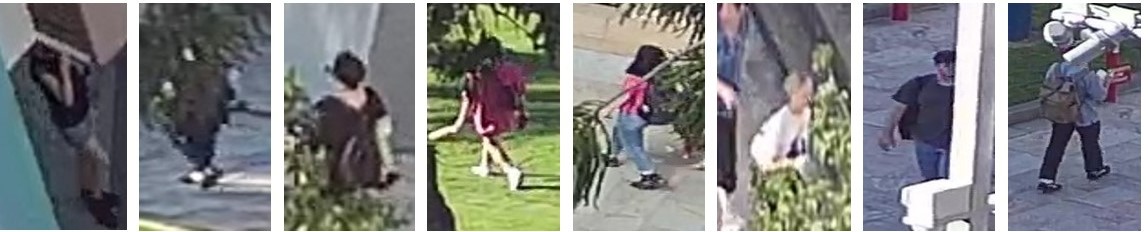} \\
		\small (c) Partially Occluded                                                  
	\end{tabular}
					  
	\begin{tabular}{@{}c@{}}
		\includegraphics[width=0.7\columnwidth, height = 1cm]{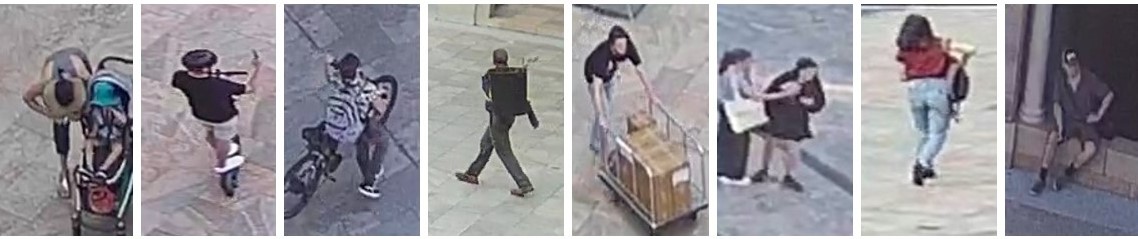} \\
		\small (d) Pose                                                  
	\end{tabular}
					  
	\begin{tabular}{@{}c@{}}
		\includegraphics[width=0.7\columnwidth, height = 1cm]{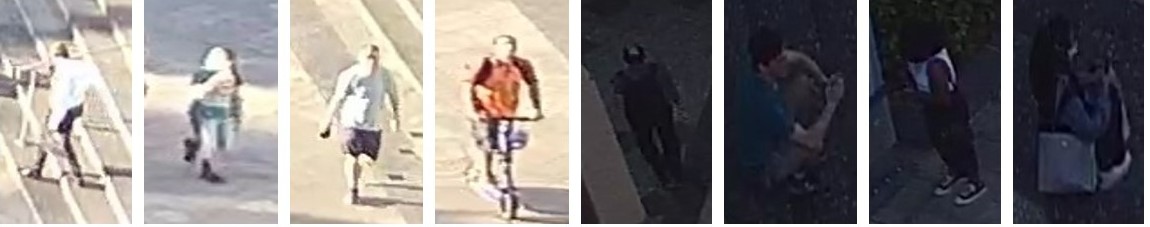} \\
		\small (e) Illumination                                                  
	\end{tabular}
					  
	\begin{tabular}{@{}c@{}}
		\includegraphics[width=0.7\columnwidth, height = 1cm]{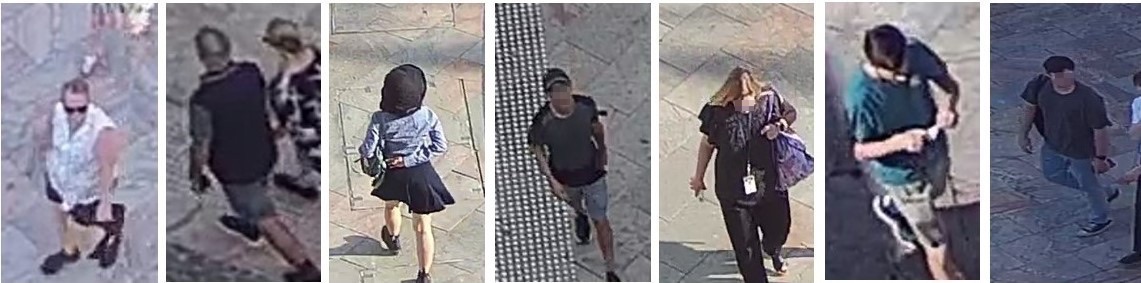} \\
		\small (f) Elevated Viewpoint                                             
	\end{tabular}
					  
	\caption{Exemplar challenges to the person re-ID task in our AG-ReID dataset due to varying flying altitudes and view variations between aerial and ground cameras.}
	\label{fig:Challenges}
\end{figure}

{In contrast to the limited setup outlined in \cite{Schumann2017PersonRA}, } AG-ReID contains data from two cameras located in heavily crowded outdoor environments as illustrated in Fig.~\ref{fig:DataCollectionArea}. The DJI XT2 drone flew at multiple altitudes ranging from 15 to 45 meters, achieving diverse viewpoints and backgrounds. In total, there are 12 pairs of videos in our dataset. We sample the video footage at 30 frames per second for both UAV camera and CCTV camera. The spatial resolution of the videos was as follows: UAV camera:  3840$\times$2160 pixels/frame;  CCTV camera: 704$\times$480 pixels/frame. The cropped person images from the UAV camera range in size from a maximum of 371$\times$678 pixels to a minimum of 31$\times$59 pixels. The cropped person images from the CCTV camera range from 172$\times$413 pixels to a minimum of 22$\times$23 pixels. 
{Compared to existing ground-ground or aerial-aerial datasets, our AG-ReID dataset presents significant challenges for person re-ID due to }the low number of pixels of the cropped person images, 
illumination changes, occlusions, motion blur, and varying poses including people walking, riding bikes or scooters, which are present in our database as illustrated in Fig. \ref{fig:Challenges}. We will provide the hard examples in the database and qualitative analysis in the document released with the database. The database was collected over multiple days under different weather conditions. Additional statistical information on the dataset will be documented with the data release. Our final dataset includes 21,983 images and 388 person identities.  
{Despite consisting of only 388 person identities, our dataset presents a statistically significant and challenging task due to the considerable variability between aerial and ground views, reflecting a practical scenario with limitations that increase the difficulty compared to simplified multi-camera configurations.}


\subsection{Annotations}

To detect tiny people in high-altitude aerial images, we use object tracking code from \cite{yolov5-strongsort-osnet-2022} to extract cropped person images from ground and aerial camera videos. Images are saved every 30 frames. We manually removed non-person images and employed three annotators to annotate the dataset, cross-checking each other's work. We annotated the AG-ReID dataset with 15 soft-biometric labels from \cite{Kumar2021ThePA} and compared its statistics with related public datasets in Table \ref{tab:compare_statics}.


\section{Two-stream Aerial-Ground Re-ID}
\begin{figure*}[hbt!]
	\centering
	\includegraphics[width=0.9\linewidth, height = 6cm]{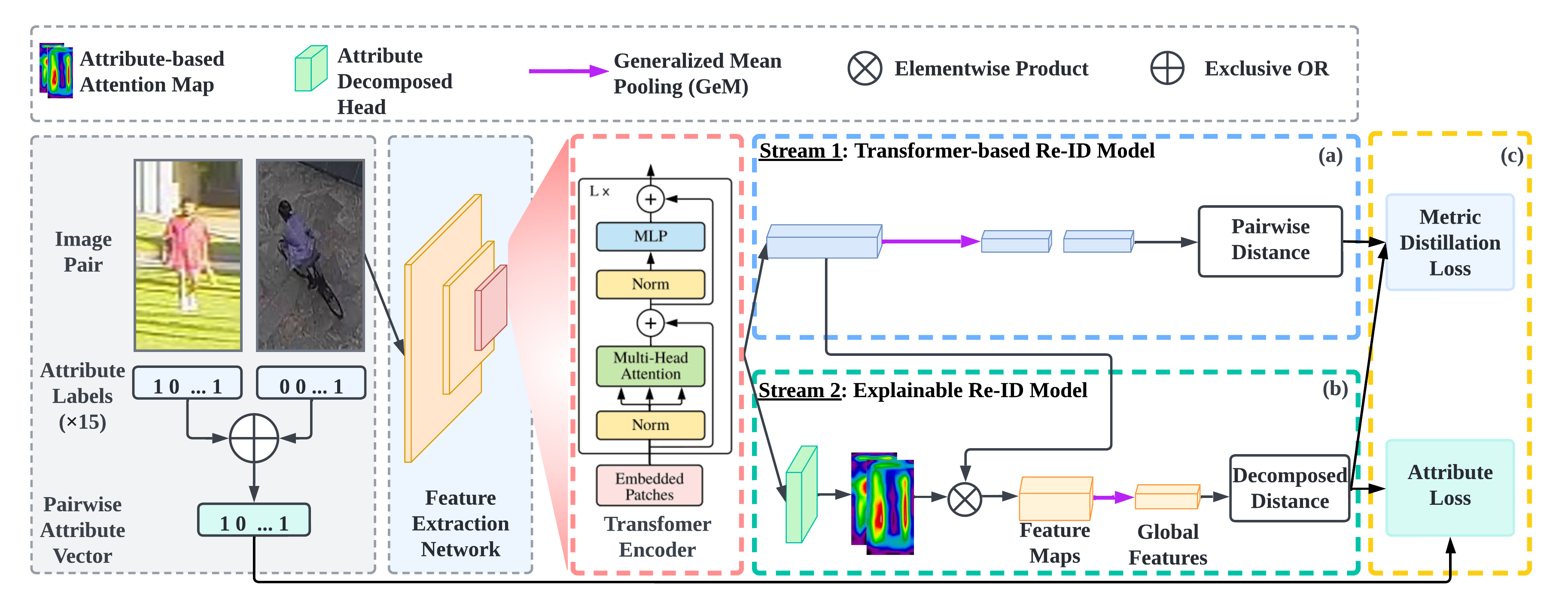}
	\caption{The overall architecture of explainable person re-ID model with vision transformer (ViT) backbone. The architecture includes two streams. While Stream 1 computes the pairwise distance from the input of pair images, Stream 2 utilizes person attributes to mitigate appearance differences of one person due to varying flying altitudes and view variations between aerial and ground cameras. Stream 2 attached with Attribute Decomposed Head (ADH) produces an attribute-guided attention map for each attribute. The two-stream model's total loss, including a metric distillation loss and an attribute loss.}
	\label{fig:explainable_overal_architecture}
\end{figure*}

The overall architecture of the proposed two-stream person re-ID model is depicted in Fig. \ref{fig:explainable_overal_architecture}. Our proposed two-stream model for the person re-ID task includes a transformer-based person re-ID stream and an explainable re-ID stream. We train the first transformer-based stream to effectively learn the discriminative feature of images and produce salient feature maps. We aggregate these feature maps with attribute attention maps generated from the explainable re-ID model trained in Stream 2 to obtain attribute feature maps. From the feature maps in Stream 1 and attribute feature maps in Stream 2, we compute two metric distances with generalized mean pooling (GeM) respectively and achieve metric distance and attribute-guided distance for our model. We describe each stream in detail as follows.

\vspace{-3px}
\subsection{Stream 1 - Transformer-based Re-ID Model}
We first train a person re-ID network, $\mathcal{F(\cdot)}$. We choose SBS \cite{He2020FastReIDAP} as the base for our re-ID model. This base is replaceable with other popular models such as MGN \cite{Wang2018LearningDF}, BOT \cite{Luo2019BagOT}, etc. Our model is built with a ViT backbone \cite{Dosovitskiy2021AnII} due to its encouraging performance and low-resolution feature maps adapting well with our dataset where aerial image's resolution is low. We also test with other backbones \eg ResNet50 \cite{He2016DeepRL}, and OSNet \cite{Zhou2019OmniScaleFL}. We can extract each image pair $(x_i, y_i)$ from the model's last layer feature map $(F_i, F_j)$. Applying generalized mean pooling (GeM), we obtain feature vector $(f_i, f_j)$ and then calculate pairwise distance $d_{i,j}$, which we use to calculate metric distillation loss in Stream 2 explainable re-ID network. The metric distillation loss is vital in keeping the distance metrics between two streams consistent.

\vspace{-6px}
\subsection{Stream 2 - Explainable Re-ID Model}
The explainable re-ID network, $\mathcal{G(\cdot)}$, has a similar structure to the re-ID model $\mathcal{F(\cdot)}$. The two models share low and mid-level layers. These layers focus on visual texture and colour, essential for learning attributes. An attributes decompose head (ADH) is designed after the last layer of explainable re-ID network $\mathcal{G(\cdot)}$. ADH consists of convolution layer $\frac{C}{8}\times3\times3$ where $C$ is the channel number of the explainable network $\mathcal{G(\cdot)} $'s last convolution layer, a $M\times1\times1$ convolution layer where $M$ is the number of attributes, and $\delta(\cdot)$ is the activation function. Activation function $\delta(\cdot)$ is important as it generates attribute-guided attention maps (AAMs) for explainable visualization and exclusive attribute learning. The sigmoid activation function $\delta(\cdot)$ is calculated as: 
\begin{equation}
	\footnotesize
	\delta(x) = \begin{cases}
	\mathcal{K}\cdot(x+1)^{\mathcal{T}}, x > 0 \\
	\mathcal{K}\cdot e^{x}, x<=0, 
	\end{cases}
	\label{eq:activate_function}
\end{equation}
where $\mathcal{K}$ is growth factor $(0,1)$ and $x$ is output of last convolution layer in ADH. With this design, ADH can effectively decompose salient regions of attributes without being biased toward the imbalanced distribution of attributes.

From ADH, we can obtain AAMs \eg $A_i$ and $A_j$ ${\in}R^{M\times w \times h}$ where $M$ is the number of attributes, $w$, and $h$ is the width and height of attention maps. Each $A_i$ and  $A_j$ contains matrix array of $M$ such as $(A_i^1, A_i^2, ..., A_i^M)$  and $(A_j^1, A_j^2, ..., A_j^M)$ where $A_i^k$ or $A_j^k$ are the attention map of $k^{th}$ attribute. Consequently, we compute attribute-guided feature map $F_i^k$ and $F_j^k$ by: 
 
\begin{equation}
	\footnotesize
	\begin{cases}
		F_i^k = F_i \bigotimes A^k  \\
		F_j^k = F_j \bigotimes A^k, 
	\end{cases}
	\label{eq:attribute_feature_map}
\end{equation}
where $A_i^k$, $A_j^k$ are attention map, $\bigotimes$ is element-wise multiplication. Each image can obtain $M$ attribute-guided feature maps where a pixel activated by attribute $k^{th}$ is highlighted, and another pixel is depressed. We then obtain attribute-guided feature vector $f_i^k$, $f_j^k$ from the attribute-guided feature map using GMP and calculate the attribute-guided distance of each attribute accordingly.  

\subsection{Loss Functions}
We calculate the loss for our explainable two-stream re-ID network. The total loss is formulated as follows:  
\begin{equation}
	\footnotesize
	L = L_d + \alpha  L_{p1} + \beta L_{p2},
	\label{eq:total_loss}
\end{equation}
where $L_d$ is metric distillation loss, $ L_{p1}$ and $L_{p2}$ are parts of attribute prior loss, and $\alpha$ and $\beta$ are balance factors.

\vspace{3px}
\noindent \textbf{Metric distillation loss.}  We calculate $L_d$ using pairwise distance $d$, the product of re-ID model $\mathcal{F(\cdot)}$ and attribute-guided distance $d_{i,j}^k$ from explainable model $\mathcal{G(\cdot)}$ as:
\begin{equation}
	\footnotesize
	L_d = |d_{i,j} - \sum^M_{k=1} d_{i,j}^k|.
	\label{eq:metric_loss}
\end{equation}

For each attribute image pair $(x_i, y_i, a_i)$ and $(x_j, y_j, a_j)$, $x_i$ is the person image, $y_i$ is the annotated identity, $a_i$ is the array of attributes $(a_i^1, a_i^2, ..., a_i^M)$ with $a_i^k$ is a binary vector, $k^{th}$ attribute and $M$ is the number of attributes, we can calculate a pairwise attribute vector, which is formulated by:
\begin{equation}
	\footnotesize
	a_{i,j} = a_i \bigoplus a_j,  
\end{equation}
where $\bigoplus$ exclusive OR and $a_{i,j}$ contains common attributes and exclusive attributes of $x_i$ and $x_j$. 
 
\vspace{3px}
\noindent \textbf{Attribute prior loss.} The equation to calculate the first part of attribute prior loss is solved by: 
\begin{equation}
	\footnotesize
	\begin{split}
		L_{p1} = max (0, (\frac{M_E}{M})^v - \sum_{e=1}^{M_E} \frac{d^e_{i,j}}{\hat{d}_{i,j}}) \\
		+ max (0, \sum ^ {M - M_e} _ {c=1} \frac{d^c_{i,j}}{\hat{d}_{i,j}} - 1 + (\frac{M_E}{M})^v ),
	\end{split}
	\label{eq:att_loss_p1}
\end{equation}
where $M_E$ is number of exclusive attributes derived from $a_{i,j}$, $M$ is number of all attributes, $v$ is vector $(0,1)$ regulating exclusive attributes, $d^e_{i,j}$ is distance of exclusive attributes, $d^c_{i,j}$ is distance of common attributes and $\hat{d}_{i,j}$ is decomposed distance from pairwise distance $d_{i,j}$,

The equation for the second part of attribute prior loss is defined as:
\begin{equation}
	\footnotesize
	\begin{split}
		L_{p2} = \sum ^{M_E} _{e=1} max (0, e^{-\lambda} \frac{(\frac{M_E}{M})^v}{M_E} - \frac{d_{i,j}^e}{\hat{d}_{i,j}}) \\
		+ \sum ^{M - M_E} _{c=1}max (0, \frac{d_{i,j}^c}{\hat{d}_{i,j}}- e^{\lambda} \frac{1-(\frac{M_E}{M})^v}{M-M_E}),
	\end{split}
	\label{eq:att_loss_p2}
\end{equation}
where
\begin{equation}
	\footnotesize
	\lambda = \frac{1}{2} ln \frac{M - M_E(\frac{M_E}{M})^v}{M_E(1-(\frac{M_E}{M})^v)}.
\end{equation}
The attribute prior loss, the sum of $ L_{p1}$ and $L_{p2}$, is the main factor in the explainable network that learns the difference between two persons through exclusive attributes effectively.
 
\section{Experiments}

\subsection{Dataset Partition}
Our AG-ReID dataset has 21,893 images and 388 identities. Initially, we had 397 identities and split them 50:50 for training and testing. We used 199 identities (11,554 images) for training and 189 identities (12,464 images) for testing, after removing nine identities due to irrelevant details. The testing set has 2,033 query images and 10,429 gallery images, with 1,701 aerial and 962 ground query images, and 7,204 aerial and 3,255 ground gallery images.

\begin{table*}
 \fontsize{8}{10}\selectfont
	\caption{Comparison of the performance of baselines and state-of-the-arts person re-ID methods on the ground-ground dataset (Market-1501), the aerial-aerial dataset (UAV-Human), and the aerial-ground dataset (AG-ReID).}
	\begin{tabularx}{\textwidth}{bssssssss}
		\hline
		\multirow{3}{*}{\textbf{Model}}
		  & \multicolumn{2}{c}{\textbf{Ground $\rightarrow$ Ground}} 
		  & \multicolumn{2}{c}{\textbf{Aerial $\rightarrow$ Aerial}} 
		  & \multicolumn{2}{c}{\textbf{Aerial $\rightarrow$ Ground}} 
		  & \multicolumn{2}{c}{\textbf{Ground$\rightarrow$ Aerial}}  \\
		  \cline{2-9}
										         
		  & \multicolumn{2}{c}{Market 1501}                          
		  & \multicolumn{2}{c}{UAV-Human}                            
		  & \multicolumn{2}{c}{AG-ReID}                              
		  & \multicolumn{2}{c}{AG-ReID}                              
		\\   \cline{2-9}
										         
		  & mAP                                           
		  & Rank-1                                        
		  & mAP                                           
		  & Rank-1                                        
		  & mAP                                           
		  & Rank-1                                        
		  & mAP                                           
		  & Rank-1                                        
		\\  \hline
										        
		BoT(R50)\cite{Luo2019BagOT}
		  & 83.95                                         
		  & 94.77                                          
		  & 63.41                                          
		  & 62.48                                         
		  & 61.37                                          
		  & 73.83                                         
		  & 62.58                                         
		  & 72.45  \\
										        
		MGN(R50)\cite{Wang2018LearningDF}
		  & 86.90                                          
		  & 95.70                                          
		  & \textbf{70.40}                                
		  & \textbf{70.38}                                
		  & 63.77                                         
		  & 75.93                                         
		  & \textbf{69.83}                               
		  & \textbf{79.42} \\
										
		SBS(R50)\cite{He2020FastReIDAP}
		  & \textbf{88.20}                               
		  & \textbf{95.40}                               
		  & 65.93                                        
		  & 66.38                                        
		  & \textbf{67.51}                               
		  & \textbf{79.85}                               
		  & 67.48                                        
		  & 77.75 \\
		\hline
	\end{tabularx}
	\label{comp_sota_dataset}
\end{table*}

\subsection{Implementation and Evaluation Metrics}
For the experiments, to make it comparable between datasets, we use state-of-the-art re-ID, \eg BoT Baseline \cite{Luo2019BagOT}, StrongerBaseline (SBS) \cite{He2020FastReIDAP}, and MGN \cite{Wang2018LearningDF}. Due to their popularity, we chose three backbones, ResNet-50, ViT (Vision Transformer) and OSNet, to build backbones for the person re-ID models. These backbones all use parameters pre-trained on ImageNet \cite{Deng2009ImageNetAL}. We train models with Adam optimizer \cite{Kingma2015AdamAM} using learning rate $lr = 10^{-4}$. For the model built with the ViT backbone, we adopt SGD optimizer with learning rate $lr = 10^{-3}$. We train the explainable network (interpreter) in Stream 2, similar to the re-ID model in Stream 1. This explainable model uses SBS \cite{He2020FastReIDAP} as the base and the identical three backbones with shared stages, i.e., convolution layers from the re-ID model during inference. We trained the explainable network on the AG-ReID training set with an attribute number configured to 88, a binary version derived from 15 annotated attributes. The explainable network is trained with Adam optimizer and a learning rate of $lr = 10^{-4}$.  For evaluation, we use two standard metrics, i.e. mean Average Precision (mAP) \cite{Zheng2015ScalablePR}, and Cumulative Matching Characteristics (CMC-$k$) or Rank-k matching accuracy \cite{Wang2007ShapeAA}. The first metric, mAP, calculates retrievals performance on average based on multiple ground truth labels. The second metric, CMC-$k$, is a probability metric measuring correct matching(s) in the retrieved results of the top-k rank. Specifically, we report rank-1, where $k=1$ for CMC-$k$.

\subsection{Comparison with State-of-the-art Models} \label{Comparison with State-of-the-art Models}

In this section, we present the performance of popular re-ID models across three datasets in Table \ref{comp_sota_dataset}. Using Strong Baseline (BoT) \cite{Luo2019BagOT} with ResNet50, we achieve 61.37\% mAP and 73.83\% rank-1 accuracy for aerial to ground, while ground to aerial shows a 1.38\% rank-1 accuracy drop and a 1.21\% mAP increase. We compare MGN \cite{Wang2018LearningDF} and Stronger Baseline (SBS) \cite{Zhou2019OmniScaleFL}  models with ResNet50 across datasets. SBS (R50) outperforms other models on our dataset with aerial to ground setting compared to the UAV-Human dataset.

\subsection{Ablation Studies}
\begin{figure}
	\centering
	\includegraphics[width=0.9\columnwidth]{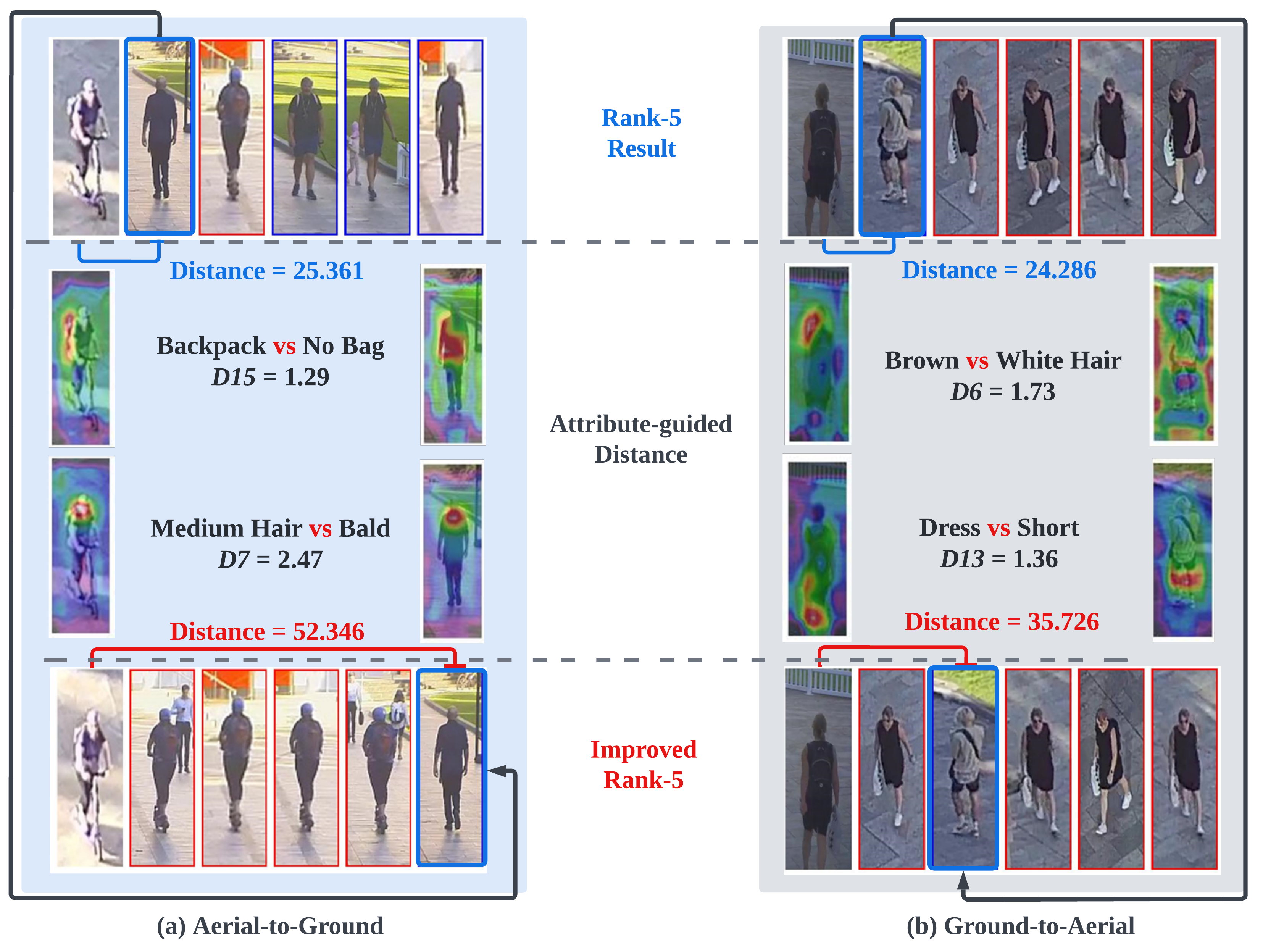}
	\caption{Visualization of the explainable re-ID model with attributes improving Rank-5 accuracy.}
	\label{fig:vis_explainable_rank1}
\end{figure}

In this study, we examined the explainable stream's effect on re-ID baselines using three backbones on the AG-ReID dataset. We observed improved rank-1 accuracy for all models in aerial-to-ground and ground-to-aerial settings, with mAP increases in two out of six models and a negligible decrease in one ($<$0.5\%). Table \ref{tab:interpreter_reweight_AG} summarizes the results. The explainable SBS model with ResNet50 backbone outperformed the re-ID model, achieving 60.93\% mAP and 74.21\% rank-1 accuracy. The ViT backbone model had a slightly lower mAP, while the OSNet model showed improvements in both mAP and rank-1 accuracy. Fig. \ref{fig:vis_explainable_rank1} illustrates how the explainable re-ID model improves rank-5 accuracy and offers attention maps for attributes impacting similarity distances.

Our baseline method shares similarities with \cite{Chen2021ExplainablePR}, but introduces innovations to improve performance by up to 7\%, setting a new benchmark on our person re-ID database. We achieved this by using 2D adaptive average pooling, automatic mixed precision (AMP) training (GradScale), and training with 88 binary vector attributes from 15 soft attribute labels. Additionally, we evaluated various backbones (ResNet, ViT, and OSNet), with ViT being optimal. The benchmark performance on this new database highlights the challenges in aerial-ground person re-ID and encourages further research in this area.

\begin{table}    
    \fontsize{8}{10}\selectfont
    \centering
    \caption{Impact of the explainable stream on the re-ID baseline with different backbones on the AG-ReID dataset.}
    \begin{tabularx}{\columnwidth}{bssss}
        \hline
        \multirow{2}{*}{\textbf{Model}}  &
        \multicolumn{2}{c}{\textbf{Aerial $\rightarrow$ Ground}} &
        \multicolumn{2}{c}{\textbf{Ground $\rightarrow$ Aerial}} \\
        \cline{2-5}
                                                     
        &
          mAP                                                       
          & Rank-1                                                
          & mAP                                                   
          & Rank-1                                                
        \\ \hline
        
        SBS(R50)
          & 59.77                                                  
          & 73.54                                                 
          & 62.27                                                 
          & 73.70                                                    
        \\ 
                                                     
        $+$ Explainable
          & 60.93                                                 
          & 74.21                                                  
          & 62.99                                                 
          & 74.74                                                    
        \\

        SBS(ViT) 
          & 67.08                                                  
          & 77.27                                                  
          & 67.14                                                    
          & 77.13                                                    
        \\ 
                                                     
        $+$ Explainable
          & 66.71                                                  
          & 77.75                                                 
          & 67.12                                                  
          & 79.52                                                    
        \\

        SBS(OSNet) 
          & 58.32                                                  
          & 72.59                                                  
          & 60.99                                                 
          & 74.22                                                  
        \\ 
                                                     
        $+$ Explainable
          & 59.73                                                     
          & 74.50                                                     
          & 62.02                                                     
          & 75.05                                                    
        \\

        BoT(R50) 
          & 55.47                                                    
          & 70.01                                                  
          & 58.83                                                     
          & 71.20                                                    
        \\ 
                                                     
        $+$ Explainable
          & 56.64                                                  
          & 71.35                                                    
          & 59.72                                                  
          & 72.45                                                    
        \\ 
        BoT(OSNet) 
          & 55.64                                                    
          & 70.29                                                  
          & 58.38                                                    
          & 73.49                                                    
        \\ 
                                                     
        $+$ Explainable
          & 56.49                                                    
          & 71.25                                                     
          & 59.15                                                 
          & 74.43                                                    
        \\ 

        BoT(ViT) 
		  & { 72.38  }                                                  
		  & {81.28  }                                                   
		  & {73.35 }                                                    
		  & { 82.64    }                                                
		\\ 
												         
		$+$ Explainable
		  & { \textbf{72.61}}                                           
		  & { \textbf{81.47}}                                           
		  & { \textbf{73.39} }                                          
		  & { \textbf{82.85} }                                          
		\\ \hline
                                       
\end{tabularx}
	\label{tab:interpreter_reweight_AG}
					 
\end{table}

\section{Conclusion}
Aerial-ground person re-ID is a growing challenge due to increased aerial surveillance systems. We introduce the first aerial-ground person re-ID dataset with 21,893 images and 388 identities, containing multiple challenges such as elevated views, low resolution, and diverse human poses. To address these, we propose an explainable re-ID approach with two streams, incorporating attributes for the aerial-ground setting. Experiments demonstrate improved ranking accuracy and performance through attribute attention maps. Our AG-ReID dataset and baseline codes are publicly available, and we plan to release video tracks to enrich the dataset further.

\section*{Acknowledgment}

This study was funded by ARC Discovery (Project No. DP200101942) and a QUT Postgraduate Research Award. Ethics approval can be provided. Participant facial regions were pixelated for privacy.

\small
\bibliographystyle{IEEEbib}
\bibliography{conference_101719_6pages}

\end{document}